%% file: icml2023.tex
\documentclass{article}

\usepackage{microtype}
\usepackage{graphicx}
\usepackage{subfigure}
\usepackage{booktabs} 

\usepackage{hyperref}

\usepackage{url}
\newcommand{\ours}{TED}
\usepackage{pifont}
\newcommand{\cmark}{\text{\ding{51}}}%
\newcommand{\xmark}{\text{\ding{55}}}%
\newcommand{\Wcal}{\mathcal{W}}
\newcommand{\Wcals}{\Wcal_{s}}
\newcommand{\Wcalt}{\Wcal_{t}}
\newcommand{\W}{W}

\newcommand{\Wsk}{\W_{s}^{k}}

\newcommand{\Wtk}{\W_{t}^{k}}

\usepackage[accepted]{icml2023}

\usepackage{amsmath}
\usepackage{amssymb}
\usepackage{mathtools}
\usepackage{amsthm}

\usepackage[capitalize,noabbrev]{cleveref}

\theoremstyle{plain}
\newtheorem{theorem}{Theorem}[section]

\theoremstyle{definition}

\theoremstyle{remark}
\newtheorem{remark}[theorem]{Remark}

\usepackage[textsize=tiny]{todonotes}

\icmltitlerunning{Less is More: Task-aware Layer-wise Distillation for Language Model Compression}

\begin{document}

\twocolumn[
\icmltitle{Less is More: Task-aware Layer-wise Distillation for Language Model Compression}



\icmlsetsymbol{equal}{*}

\begin{icmlauthorlist}
\icmlauthor{Chen Liang}{yyy}
\icmlauthor{Simiao Zuo}{comp}
\icmlauthor{Qingru Zhang}{yyy}
\icmlauthor{Pengcheng He}{comp}
\icmlauthor{Weizhu Chen}{comp}
\icmlauthor{Tuo Zhao}{yyy}
\end{icmlauthorlist}

\icmlaffiliation{yyy}{H. Milton Stewart School of Industrial and Systems Engineering, Georgia Institute of Technology, Atlanta, U.S.A.}
\icmlaffiliation{comp}{Microsoft, Redmond, U.S.A.}

\icmlcorrespondingauthor{Chen Liang}{cliang73@gatech.edu}
\icmlkeywords{Machine Learning, ICML}

\vskip 0.3in
]



\printAffiliationsAndNotice{}  

\input{01-abstract}
\input{02-introduction}

\input{03-background}
\input{04-method}
\input{05-experiment}
\input{06-analysis}

\input{07-discussion}
\input{08-conclusion}

\bibliography{icml2023}
\bibliographystyle{icml2023}

\newpage
\appendix
\onecolumn
\input{appendix}

\end{document}

%% file: 01-abstract.tex
\begin{abstract}

Layer-wise distillation is a powerful tool to compress large models (i.e. teacher models) into small ones (i.e., student models). The student distills knowledge from the teacher by mimicking the hidden representations of the teacher at every intermediate layer. However, layer-wise distillation is difficult. Since the student has a smaller model capacity than the teacher, it is often under-fitted. Furthermore, the hidden representations of the teacher contain redundant information that the student does not necessarily need for the target task's learning. To address these challenges, we propose a novel Task-aware layEr-wise Distillation (TED). 
TED designs task-aware filters to align the hidden representations of the student and the teacher at each layer. 
The filters select the knowledge that is useful for the target task from the hidden representations. As such, TED reduces the knowledge gap between the two models and helps the student to fit better on the target task.
We evaluate TED in two scenarios: continual pre-training and fine-tuning. TED demonstrates significant and consistent improvements over existing distillation methods in both scenarios. Code is available at \url{https://github.com/cliang1453/task-aware-distillation}.


\end{abstract}


%% file: 02-introduction.tex
\section{Introduction}

Large pre-trained language models have achieved state-of-the-art performances in many natural language processing tasks \citep{wang2018glue,rajpurkar2016squad}. However, their deployment in resource-limited scenarios is hindered by their huge number of parameters \citep{raffel2019exploring,radford2019language,brown2020language,he2020deberta,he2023debertav}. Knowledge Distillation (KD) \citep{hinton2015distilling} is a powerful tool to compress large models (i.e., teacher models) into small ones (i.e., student models) with a minimal loss of performance. This approach trains the student to match the output predictions of the teacher.

However, such a last-layer-only distillation approach does not exploit the intermediate layers of the teacher, which contain rich semantic and syntactic knowledge. To leverage such knowledge, researchers have proposed a \textit{layer-wise distillation} approach, which trains the student to match the hidden representation of the teacher at each layer \citep{sun2019patient,jiao2019tinybert,sun2020mobilebert,hou2020dynabert,zuo2022moebert}. Such an approach often improves the generalization performance of the student model.

Nevertheless, layer-wise distillation faces two major challenges. First, the student may struggle to mimic the hidden representations of the teacher due to their large capacity gap. This often leads to large discrepancies between their hidden representations. Consequently, model training/optimization often favors reducing such large discrepancies over the training loss of the student (i.e., the target task's loss such as cross-entropy), resulting in an under-fitted student model. Second, mimicking the hidden representations may not be beneficial for the target task's learning. This is because the hidden representations of the teacher often contain redundant information \citep{dalvi2020analyzing,durrani2020analyzing}. Given the limited capacity of the student, such redundant information may compete with the useful information for distillation, hindering the useful knowledge from being distilled. Our empirical observations show that for some tasks, layer-wise distillation only marginally outperforms standard KD (Table~\ref{tb:deberta_glue}).

\begin{figure*}[t!]
    \centering
    \includegraphics[width=0.9\linewidth]{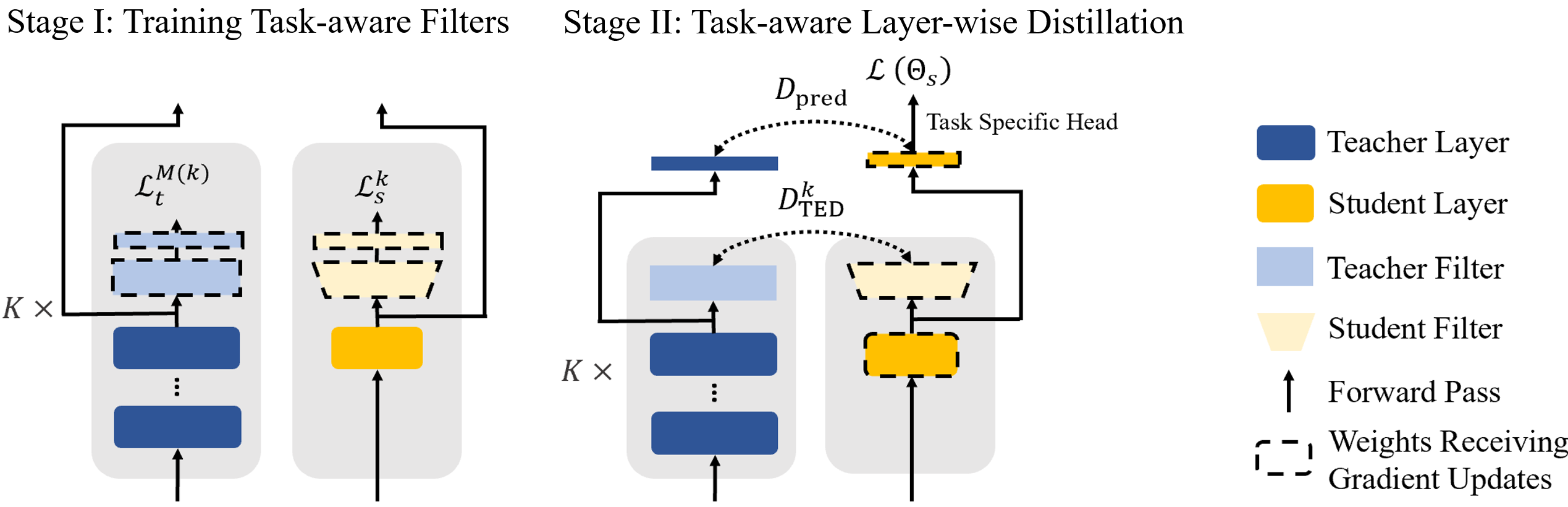}
    \caption{
    An illustration of {\ours}'s two-stage training framework. In \textit{Stage I} (left), we fix the model parameters and only train the filters and task-specific heads based on the target task loss. In \textit{Stage II} (right), we jointly train the student and its filters by aligning the filter outputs of each pair of the teacher and the student layers.
    }
	\label{fig:framework}
\end{figure*}

To address these challenges, we propose a novel layer-wise distillation method, {\ours} (\underline{\textbf{T}}ask-aware lay\underline{\textbf{E}}r-wise \underline{\textbf{D}}istillation), which distills task-specific knowledge from the teacher to the student. We design a pair of task-aware filters for each layer of the teacher and student\footnote{For simplicity, we assume that the student and teacher are of the same depth (number of layers) but different widths. The case of different depths will be elaborated in Section~\ref{sec:background}.}. Each filter is a neural network with a task-specific head (e.g., a linear soft-max layer for classification), and is trained to extract the predictive knowledge from the hidden representation of the corresponding model. Figure~\ref{fig:framework} illustrates the training procedure of {\ours}, which consists of two stages:

\noindent $\bullet$ Stage I: We train the task-aware filters for both the teacher and the student models, while keeping the model parameters frozen. At each layer, the filter takes the hidden representation of the model as input, and produces a target task's loss (e.g., cross-entropy) as output. The filter is subsequently optimized based on such a loss to capture the predictive knowledge from the hidden representation.

\noindent $\bullet$ Stage II: We jointly train the student model and its task-aware filters, while keeping the teacher and its filters fixed. At each layer, we feed the hidden representation of the teacher and the student to their respective filters (without the task-specific heads). Then, we adopt a regularizer that penalizes the discrepancy between the filtered representations. This regularizer encourages the student to learn the task-specific knowledge from the teacher, while ignoring the redundant information.

The task-aware filters serve as a selection mechanism that reduces the knowledge gap between the teacher and the student and encourages the distillation of task-specific knowledge. This makes distillation easier for the student. 

We evaluate {\ours} on two settings: continual pre-training and task-specific fine-tuning. In the continual pre-training setting, we distill a $6$-layer GPT-2 student model ($82$M) from a $12$-layer GPT-2 teacher model ($125$M) \citep{radford2019language}. We show that {\ours} outperforms existing methods in both zero-shot and transfer learning settings on various downstream tasks \citep{paperno2016lambada, merity2016pointer}. In the task-specific fine-tuning setting, we distill a DeBERTaV3-xsmall student model ($70$M) from a DeBERTaV3-base teacher model ($183$M) \citep{he2023debertav}. We demonstrate that {\ours} achieves significant improvement on the GLUE benchmark \citep{wang2018glue} and the SQuAD v1.1/2.0 question answering datasets \citep{rajpurkar2016squad, rajpurkar2018know}.

The rest of the paper is organized as follows: Section~\ref{sec:background} briefly reviews the background; Section~\ref{sec:method} presents our proposed method; Section~\ref{sec:lm} presents experiments on language modeling; Section~\ref{sec:nlu} presents experiments on natural language understanding; Section~\ref{sec:analysis} presents analysis of models; and Section~\ref{sec:conclusion} discusses and concludes the paper.

%% file: 03-background.tex
\section{Background}
\label{sec:background}

{\bf Transformer-based Language Models}. The Transformer architecture is a powerful neural network design for modeling sequential data, such as natural language \citep{vaswani2017attention,devlin2018bert,radford2019language,he2023debertav}. It consists of multiple layers that are stacked on top of each other. Each layer performs two operations: a multi-head self-attention mechanism and a two-layer feed-forward neural network. We use $f(\cdot; \Theta)$ to denote a Transformer-based model $f$ that has a set of parameters $\Theta$, where $f$ takes an input sequence $x$ from the input sample space $\mathcal{X}$ and produces an output prediction. 
We define the loss function $\mathcal{L}(\Theta) = \mathbb{E}_{x\sim \mathcal{X}}[\ell(f(x; \Theta))]$, where $\ell$ is the target task loss. For example, $\ell$ is the causal language modeling loss for generative models (i.e., $\sum_{t=1}^{|x|} \log p(x_t|x_{<t}; \Theta)$).

{\bf Knowledge Distillation} is a powerful approach to compress large models (i.e., teacher models) into smaller models (i.e., student models) by transferring knowledge from the former to the latter \citep{hinton2015distilling}. The student is trained to mimic the output predictions of the teacher. Specifically, we denote the teacher as $f_t(\Theta_t)$ and the student as $f_s(\Theta_s)$ and consider the following optimization problem:
\begin{align}
    \min_{\Theta_s} \mathcal{L}(\Theta_s) + \mathcal{D}_{\rm pred}(\Theta_t, \Theta_s),
    \label{eq:general_loss}
\end{align}
where $\mathcal{D}_{\rm pred}(\Theta_t, \Theta_s)$ is the distillation loss, a distance metric between the output predictions of the teacher and the student. For example, $\mathcal{D}_{\rm pred}$ can be the KL-divergence: $\texttt{KL}(f_t(\Theta_t)/T, f_s(\Theta_s)/T)$,
where $T>0$ is the temperature that controls the softness of the prediction probability distributions \citep{hinton2015distilling}. A commonly adopted distillation scheme is the offline distillation, where the teacher is fully-trained and fixed, and the student is optimized based on Eq.~\ref{eq:general_loss}.

{\bf Layer-wise Distillation}. In large Transformer-based models, the output predictions of the models may not capture all the semantic and syntactic knowledge encoded in the intermediate layers. Therefore, researchers propose a layer-wise distillation approach, which aligns the hidden representations of the student and the teacher at each layer \citep{romero2014fitnets,sun2019patient,sun2020mobilebert,jiao2019tinybert,hou2020dynabert,zuo2022moebert,liang2023homodistil}. 
Specifically, we denote the hidden representation at the $k$-th layer of a $K$-layer student as $H_s^k \in \mathbb{R}^{|x| \times d_s}$, and at the $M(k)$-th layer of the teacher as $H_t^{M(k)} \in \mathbb{R}^{|x| \times d_t}$. Here $|x|$ is the sequence length; $d_s$ and $d_t$ are the hidden dimensions of the student and the teacher, respectively. $M(\cdot)$ is a layer mapping function that determines from which layer in the teacher that a student layer should distill. For example, if we set $M(k) = 2k$, the student would distill from every other layer in the teacher. The layer-wise distillation loss is defined as:
\begin{align}
\mathcal{D}_{\rm layer}(\Theta_t, [\Theta_s, \Wcals]) = \sum_{k=1}^K \texttt{MSE}(H_t^{M(k)}, H_s^k \Wsk).
\label{eq:hn_loss}
\end{align}
Here $\texttt{MSE}(\cdot, \cdot)$ is the mean-squared error, $\Wsk \in \mathbb{R}^{d_s \times d_t}$ is a randomly initialized and learnable linear projection that projects $H_s^k$ into the same space as $H_t^{M(k)}$, and $\Wcals = \{\Wsk \}_{k=1}^K$.
In practice, the student is often optimized using multiple distillation losses, e.g.,
\begin{align}
\min_{\Theta_s,\Wcals} \mathcal{L}(\Theta_s) + \alpha_1 \mathcal{D}_{\rm pred}(\Theta_t, \Theta_s) + \alpha_2 \mathcal{D}_{\rm layer}(\Theta_t, [\Theta_s,\Wcals]).
\label{eq:total_loss}
\end{align}
where $\alpha_1, \alpha_2 \geq 0$ are hyper-parameters. Besides the intermediate layers, distilling knowledge from the attention scores and the embedding layers can also improve the distillation performance \citep{sun2020mobilebert, jiao2019tinybert, wang2020minilm, wang2020minilmv2}. Eq.~\ref{eq:total_loss} can be further extended by adding such losses.

%% file: 04-method.tex
\section{Method}
\label{sec:method}

We introduce {\ours}, a two-stage training framework that uses task-aware filters to distill knowledge from a teacher to a student. The task-aware filters are neural networks that learn to extract task-specific knowledge from the hidden representations of the teacher and the student. In the first stage, we add a task-aware filter to each layer of the teacher and the student. We train these filters using the task-specific loss while keeping the model parameters frozen. In the second stage, we fine-tune the student and its filters by minimizing the discrepancy between the filtered representations of the teacher and the student. 

\subsection{Stage I: Training Task-aware Filters}

For a student that contains $K$ layers, we select $K$ corresponding layers from the teacher to match with the student using a layer mapping function, $M(\cdot)$, as defined in Section~\ref{sec:background}. We then equip each layer with a task-aware filter to extract the task-specific knowledge from the hidden representation of this layer. Each filter is a neural network with a task-specific head (e.g., a linear soft-max layer for classification). It takes in the hidden representation generated by this layer and outputs a prediction for the target task. For example, for a classification task, the filter outputs a probability distribution over the classes.

For simplicity, we only specify how to train task-aware filters for the teacher. The student is treated similarly (see Section~\ref{sec:lm} for details). To train the task-aware filters, we fix the parameters of the teacher, which is already pre-trained \footnote{We discuss in detail how to initialize the teacher and the student models in Section~\ref{sec:lm} and \ref{sec:nlu}.}. In other words, we only update the parameters of the filters. We denote the task-aware filter at the $M(k)$-th layer as $g_t^{k}(\cdot ; \Wtk)$, where $\Wtk$ is the filter's parameters. The filter takes in the hidden representation $H_t^{{M(k)}}$ at the $M(k)$-th layer, and outputs a task-specific loss
\begin{align}
    \mathcal{L}_t^{k} (\Theta_t^{M(k)}, \Wtk)
    = \mathbb{E}_{x\sim\mathcal{X}}[\ell(g_t^{k}(H_t^{{M(k)}}; \Wtk ))], 
    \label{eq:teacher_filter_loss}
\end{align}
where $\Theta_t^{M(k)}$ is the teacher's parameters up to the $M(k)$-th layer.
The loss function $\ell$ depends on the task and the setting. For example, $\ell$ is the causal language modeling loss for continual pre-training and the cross-entropy loss for fine-tuning of classification tasks. Given the loss in Eq.~\ref{eq:teacher_filter_loss}, we train the $K$ filters jointly:
\begin{align}
    \min_{\Wcalt} \ \sum_{k=1}^K \mathcal{L}_t^{k} (\Theta_t^{M(k)}, \Wtk),
    \label{eq:ted_filters_loss}
\end{align}
where $\Wcalt = \{W_t^k\}_{k=1}^K$. By training the task-aware filters, we can reduce the redundant information in the hidden representations, and keep the information that is useful for learning the target task.

\remark We can choose different neural network architectures to implement the task-aware filters, such as a simple linear projection that maps the input to a lower-dimensional space, a multi-layer perceptron that applies a sequence of nonlinear transformations, or a stack of Transformer layers that encode the input with attention mechanism. We compare the performances of these architectures in Section~\ref{ana:complexity}.

\subsection{Stage II: task-aware Layer-wise Distillation}

In Stage II, we remove the task-specific heads in the task-aware filters, which are learned in Stage I. Then, we freeze the parameters of the teacher and its filters, and fine-tune the student and its filters by minimizing the discrepancy between the filtered representations at each layer of the two models.

Formally, we denote $g_s^k(\cdot, \Wsk)$ as the task-aware filters at the $k$-th layer of the student. Then the task-aware layer-wise distillation loss is defined as
\begin{align}
    \mathcal{D}_{\rm \ours}\left( \left[\Theta_t, \Wcalt\right], \left[\Theta_s, \Wcals\right]\right) \quad\quad\quad\quad\quad\quad\quad\quad\quad\quad \nonumber\\ = \sum_{k=1}^K  \texttt{MSE}\left(g_t^{k}(H_t^{{M(k)}}; \Wtk), g_s^{k}(H_s^{k}; \Wsk)\right),
    \label{eq:ted_hn_loss}
\end{align}
which measures the discrepancy between the filtered representations of the teacher and the student. Based on the distillation loss, the training objective for the student and its filters is
\begin{align} 
    \min_{\Theta_s, \Wcals} \mathcal{L}(\Theta_s) + \alpha_1 \mathcal{D}_{\rm pred}(\Theta_t, \Theta_s) \quad\quad\quad\quad\nonumber\\+ \alpha_2 \mathcal{D}_{\rm \ours}([\Theta_t, \Wcalt],[\Theta_s, \Wcals]),
    \label{eq:ted-total}
\end{align}
where $\mathcal{L}$ is the target task's loss and $\mathcal{D}_{\rm pred}$ is the prediction distillation loss defined in Eq~\ref{eq:general_loss} and $\alpha_1, \alpha_2 \geq 0$ are hyper-parameters. By using the task-aware filters, Eq.~\ref{eq:ted-total} imposes an easier requirement on the student than the conventional layer-wise distillation loss (Eq.~\ref{eq:total_loss}). That is, Eq.~\ref{eq:total_loss} requires the student to match the teacher on the unfiltered hidden representations, regardless of their relevance to the target task.

\remark We can also keep the task-specific heads in the task-aware filters and penalize the KL-divergence instead of the mean-squared error. We compare the performances of these two variants in Section~\ref{ana:feature_type}.

%% file: 05-experiment.tex
\section{Language Modeling}
\label{sec:lm}

\subsection{Data}

First, we evaluate {\ours} in the continual pre-training setting by distilling generative models on language modeling tasks. We use \textbf{Open WebText}\footnote{https://huggingface.co/datasets/openwebtext} \citep{gokaslan2019openwebtext}, an open-source replication of the OpenAI WebText corpus \citep{radford2019language} for open domain pre-training. It is a massive English corpus containing $8$M training documents and around $38$GB of texts extracted from $45$M links of Reddit post urls. Data pre-processing details are deferred to Appendix~\ref{app:gpt_data}.

Second, we evaluate the distilled student model by conducting zero-shot and transfer learning experiments on two downstream tasks: \textbf{LAMBADA} \citep{paperno2016lambada} and \textbf{WikiText-103} \citep{merity2016pointer}. LAMBADA evaluates the ability of language models in modeling long-range dependencies. The dataset consists of full texts of $2662$ novels extracted from BookCorpus \citep{zhu2015aligning}. WikiText-103 is a collection of over $100$M tokens extracted from the set of verified good and featured articles on Wikipedia.

\subsection{Models} 

\textbf{Teacher Model.} We use a pre-trained GPT-2 \citep{radford2019language} as the teacher model. GPT-2 is a Transformer-based model trained on Open WebText using a causal language modeling objective. We adopt the base version of GPT-2 (GPT-2$_{12}$, $125$M parameters), which contains $12$ layers and has a hidden dimension of $d_t = 768$.

\textbf{Student Model.} We initialize a $6$-layer (i.e., $K=6$) student model (GPT-2$_6$, $82$M parameters) with a subset of layers from the teacher. We adopt the layer mapping function $M(k) = 2k-1$ for $k \leq K/2$ and $M(k) = 2k$ for $k > K/2$ following \citealt{sanh2019distilbert}. We further discuss how to initialize the student model when its architecture is not a shallow version of the teacher in Appendix~\ref{app:initialization}.

\begin{table*}[t!]
\centering
\caption{Zero-shot and transfer learning performance of GPT-2$_{6}$ models on test sets. We report the results of DistilGPT-2 from \citet{sanh2019distilbert}, and the results of GPT-2$_{12}$ from \citet{radford2019language}. Other results are from our own implementation.}
\vspace*{1mm}
\resizebox{0.95\textwidth}{!}{
\begin{tabular}{l|c|ccc|ccc}
\toprule
Method &       Test     & \multicolumn{3}{c|}{Zero-Shot}    & \multicolumn{3}{c}{Transfer Learning} \\
& Open WebText  & WikiText-103    & \multicolumn{2}{c|}{LAMBADA} & WikiText-103     & \multicolumn{2}{c}{LAMBADA}     \\
& ppl$\downarrow$           & ppl$\downarrow$             & ppl$\downarrow$          & Acc$\uparrow$         & ppl$\downarrow$            & ppl$\downarrow$    & Acc$\uparrow$        \\ \midrule
GPT-2$_{12}$ (Teacher)               & 23.1         & 37.5           & 35.1          &46.0          & 15.9            & 37.2   & 34.8         \\ \midrule
DistilGPT-2$_6$ (KD)                 & 31.9         & -               & -            & -            & 21.1            & -      & -            \\  
DistilGPT-2$_6$ (KD, Re-Imp)          & 29.1         & 49.0           & 87.9          &22.9          & 19.3            & 50.1   & 31.7        \\
GPT-2$_6$ (LWD)                      & 29.7         & 51.9           & 91.9          &22.0          & 19.3            & 50.6   & 31.5   \\ \midrule
GPT-2$_6$ (\ours)                    & \textbf{28.5}         & \textbf{48.1}           & \textbf{87.2}          &\textbf{23.0}          & \textbf{19.0}            & \textbf{48.6}   & \textbf{32.1}   \\ \bottomrule
\end{tabular}}
\label{tb:gpt_main}
\end{table*}

\subsection{Training}
\label{sec:gpt_filter}

\textbf{Stage I.} For the teacher model, we design each filter as a linear projection, i.e., $\Wtk \in \mathbb{R}^{d_t \times d_t}$, and randomly initialize a filter for each layer that is selected to match with a student layer. We fix the parameters of the teacher model and train the filters based on Eq~\ref{eq:ted_filters_loss} for one epoch. We use AdamW \citep{loshchilov2017decoupled} as the optimizer and use $4$k tokens as the batch size. We adopt a linear decay learning rate schedule with a learning rate of $2.5\times 10^{-4}$ and a warmup ratio of $0.05$. Then, we directly take the trained filter at the $M(k)$-th layer of the teacher to be the filter at the $k$-th layer of the student without further training. It is intuitive that the trained filters of the teacher can serve as sufficiently good filters of the student because the student is initialized with a subset of layers from the teacher. Full implementation details are deferred to Appendix~\ref{app:gpt_filter}. 

\textbf{Stage II.} We train the student and its filters based on Eq~\ref{eq:ted-total} for four epochs. We follow the same hyper-parameter configurations as in Stage I, and set $\alpha_1 = 2.5$, $\alpha_2 = 0.1$ and temperature $T=2.0$. 

\textbf{Baselines.} We consider two baseline methods: 1) \textbf{KD} optimizes the student model based on $\mathcal{L}(\Theta_s) + \alpha_1 \mathcal{D}_{\rm pred}(\Theta_t, \Theta_s)$ (Eq.~\ref{eq:general_loss}), which is adopted by DistilGPT-2$_{6}$ \citep{sanh2019distilbert}. 2) \textbf{LWD} optimizes the student model based on $\mathcal{L}(\Theta_s) + \alpha_1 \mathcal{D}_{\rm pred}(\Theta_t, \Theta_s) + \alpha_2 \mathcal{D}_{\rm layer}(\Theta_t, [\Theta_s,\Wcals])$ (Eq.~\ref{eq:total_loss}). 

\subsection{Main Results}

Table~\ref{tb:gpt_main} shows the zero-shot and transfer learning performance of the GPT-2$_6$ models. For Open WebText, we split $5$\% for testing. For the zero-shot setting, we directly evaluate the student model on the test sets. For the transfer learning setting, we fine-tune the student model on the downstream language modeling tasks. 
We have the following observations: 1) LWD does not always lead to a better performance than KD, suggesting that the student may have difficulty mimicking the teacher at every layer. 2) {\ours} can significantly improve model performance, especially on Open WebText. This suggests that distilling the task-specific knowledge to the student yields a better model.

\begin{table*}[t!]
\centering 
\caption{Evaluation results on GLUE dev set. The teacher is a fine-tuned DeBERTaV3-base model ($183$M) and the student is a DeBERTaV3-xsmall model ($70$M). Results of ``Fine-tune'' are obtained by directly fine-tuning the DeBERTaV3-xsmall model on the target task without distillation.}
\vspace*{1mm}
\resizebox{0.95\textwidth}{!}{
\begin{tabular}{l|ccccccccccc}
\toprule
Method & MNLI-m/mm & QQP & QNLI & SST-2 & RTE & CoLA & MRPC & STSB & Avg. \\
& Acc & Acc/F1 & Acc & Acc & Acc & Mcc & Acc/F1 & P/S & Score    \\\midrule
Teacher\textsubscript{base} & 90.5/90.6 & 92.3/89.7 & 94.2 & 96.0  & 86.1 & 68.8 & 90.8/93.5 & 92.4/92.2 & 88.9 \\\midrule
Fine-tune\textsubscript{xs}  & 88.3/88.1 & 91.7/88.8 & 92.5 & 93.5  & 79.7 & 68.3 & 90.2/93.0 & 90.9/90.5 & 86.9 \\
KD\textsubscript{xs} & 88.5/88.1 & 91.7/88.8 & 92.9 & 93.9  & 80.5 & 66.3 & \textbf{91.2}/\textbf{93.7} & 91.0/90.8 & 87.0 \\
LWD\textsubscript{xs} & \textbf{88.8}/88.3 & 91.8/89.0 & 92.9 & 93.9  & 80.2 & 66.8 & 90.2/93.0 & 91.0/90.6 & 87.0 \\\midrule 
{\ours}\textsubscript{xs} & \textbf{88.8}/\textbf{88.7} & \textbf{92.2}/\textbf{89.5} & \textbf{93.1} & \textbf{94.2} & \textbf{81.8} & \textbf{68.5} & 90.4/93.2 & \textbf{91.3}/\textbf{91.1} & \textbf{87.5} \\\bottomrule
\end{tabular}
}
\label{tb:deberta_glue}
\end{table*}

\begin{table*}[t!]
\centering 
\caption{Evaluation results on SQuAD v1.1 and SQuAD v2.0 validation sets. The teacher is a fine-tuned DeBERTaV3-base model and the student is a DeBERTaV3-xsmall model.}
\vspace*{1mm}
\label{tb:deberta_squad}
\resizebox{0.48\textwidth}{!}{
\begin{tabular}{l|ccc}
\toprule
Method& SQuAD v1.1 & SQuAD v2.0 & Avg. \\ %
& EM/F1 & EM/F1 & Score\\ \midrule %
Teacher\textsubscript{base} & 87.1/93.1 & 85.4/88.4 & 90.8 \\ \midrule%
Fine-tune\textsubscript{xs} & 83.5/90.4 & 82.0/84.8 & 87.6 \\  %
KD\textsubscript{xs}        & 84.8/91.4 & 82.6/85.5 & 88.5 \\ %
LWD\textsubscript{xs}       & 84.9/91.5 & 82.8/85.6 & 88.6 \\ \midrule %
{\ours}\textsubscript{xs} & \textbf{85.4}/\textbf{91.7} & \textbf{83.0}/\textbf{85.8} & \textbf{88.8} \\ \bottomrule %
\end{tabular}}
\end{table*}

\section{Natural Language Understanding}
\label{sec:nlu}

\subsection{Data} 

We further evaluate {\ours} on natural language understanding (NLU) tasks. We consider the widely used General Language Understanding Evaluation (\textbf{GLUE},~\citealt{wang2018glue}) benchmark, which contains nine NLU tasks, including textual entailment, sentiment analysis and text similarity. We also evaluate {\ours} on \textbf{SQuAD v1.1/2.0} \citep{rajpurkar2016squad, rajpurkar2018know}, which are widely used question answering datasets. Details about the datasets are deferred to Appendix~\ref{app:nlu_data}.

\subsection{Models}
\label{sec:nlu_model}

We use DeBERTaV3 models \citep{he2023debertav} as the teacher and student models. DeBERTaV3 is pre-trained in an ELECTRA-style \citep{clark2020electra} on $160$GB open-domain corpus \citep{gokaslan2019openwebtext,trinh2018simple,nagel2016cc}. It improves BERT \citep{devlin2018bert} with disentangled attention and enhanced mask decoder, and achieves the state-of-the-art downstream performance.

\textbf{Teacher Model.} We initialize the teacher model for each task with a DeBERTaV3-base model that has been fine-tuned on the target task. The model has $183$M parameters, $12$ layers and a hidden dimension of $768$ (i.e., $d_t = 768$). We fine-tune the model using AdamW as the optimizer. We adopts a linear decay learning rate schedule with a warmup ratio in $\{0.05, 0.1\}$. We choose the learning rate from $\{1, 1.5, 2, 2.5, 3\} \times 10^{-5}$, the batch size from $\{16, 32, 64\}$, the number of training epochs from $\{3,6,8\}$ and the dropout ratio from $\{0.05, 0.1\}$. Full implementations details are deferred to Appendix~\ref{app:nlu_model}.

\textbf{Student Model.} We initialize the student model for each task with a DeBERTaV3-xsmall model that has been fine-tuned on the target task. The model contains a total $70$M parameters (including $22$M backbone parameters). It has $12$ layers and a hidden dimension of $384$ (i.e., $d_s = 384$). We further discuss how to initialize the student model when there does not exist a pre-trained or fine-tuned model with the desired architecture in Appendix~\ref{app:initialization}.

\subsection{Training}

\textbf{Stage I.} Since the teacher and student have different hidden dimensions, we set $\Wtk \in \mathbb{R}^{d_t\times d_t}$ and $\Wsk \in \mathbb{R}^{d_s\times d_t}$. We randomly initialize a filter for each layer of the student and the teacher (recall that they have the same number of layers). We freeze the model parameters of the teacher and the student, and train their filters on the target task following the same hyper-parameter configurations in Section~\ref{sec:nlu_model}. Full implementations details are deferred to Appendix~\ref{app:nlu_training}.

\textbf{Stage II.} We then train the student and its filters based on Eq~\ref{eq:ted-total} on the target task. We follow the same hyper-parameter configurations as in Stage I. We choose $\alpha_1 \in \{1.0, 2.5, 5.0, 10.0\}$, choose $\alpha_2 \in \{10, 20, 50, 100, 200, 500, 1000\}$, and set the temperature $T=2.0$. 

\subsection{Main Results}

Table~\ref{tb:deberta_glue} and Table~\ref{tb:deberta_squad} show the evaluation results of the student on the GLUE benchmark and SQuAD v1.1/2.0 datasets, respectively. {\ours} achieves consistent and significant gains over nine out of ten tasks over the best distillation baseline. For example, {\ours} achieves a gain of $0.5$ on some large datasets, e.g., QQP, and a gain of $1.0$ on some small datasets, e.g., RTE. For certain small datasets (e.g., RTE, MRPC, STS-B), LWD does not always produce a better performance than KD. In contrast, {\ours} improves upon KD in two out of three cases.

\subsection{BERT Experiments}

To compare with the state-of-the-art task-specific distillation baselines, we present in Table~\ref{tb:bert_glue} the results of a $6$-layer BERT-base student model (BERT-base$_6$, $66$M) distilled from a fine-tuned $12$-layer BERT-base teacher model (BERT-base$_{12}$, $109$M). {\ours} achieves comparable performance with noticeable benefits over the existing baselines on three NLU tasks. All implementation details are deferred to Appendix~\ref{app:nlu_bert}. 

\begin{table}[t!]
\centering 
\caption{Evaluation results on GLUE dev set. The teacher is a fine-tuned BERT-base$_{12}$ ($12$ layers), and the student is a BERT-base$_{6}$ ($6$ layers), except for CoDIR, which uses a RoBERTa-base as the teacher. References: PKD \citep{sun2019patient}, BERT-of-Thesus \citep{xu2020bert}, MixKD \citep{liang2020mixkd}, ProKT \citep{shi2021follow}, CoDIR \citep{sun2020contrastive}.}
\vspace{1.5mm}
\label{tb:bert_glue}
\resizebox{0.40\textwidth}{!}{
\begin{tabular}{l|ccc}
\toprule
Method & MNLI-m/mm & SST-2 & RTE\\ \midrule
Teacher\textsubscript{12} & 84.5/84.7 &  92.6 & 71.3 \\\midrule
KD\textsubscript{6} & 82.1/82.3 & 90.8 & 65.4 \\
LWD\textsubscript{6} & 82.7/83.1 & 90.9 & 67.6 \\ 
PKD\textsubscript{6} & 81.3/- & 91.3 & 66.5 \\
Thesus\textsubscript{6} & 82.3/- & 91.5 & 68.2 \\
MixKD\textsubscript{6}  & 82.5/- & 92.1 & 67.9 \\
ProKT\textsubscript{6} & 82.8/83.2 & 91.3 & 68.4 \\
CoDIR\textsubscript{6} & \textbf{83.6}/82.8 & \textbf{93.6} & 65.6 \\\midrule
{\ours}\textsubscript{6} & 83.4/\textbf{84.0} & 91.7 & \textbf{68.8} \\
\bottomrule
\end{tabular}}
\end{table}

%% file: 06-analysis.tex
\section{Analysis}
\label{sec:analysis}

We further verify that the task-aware filters can capture the task-specific knowledge and ease distillation. All implementation details are deferred to Appendix~\ref{app:ana}.

\subsection{Filters Capture Task-Specific Knowledge}

\begin{table*}[htb!]
\centering
\caption{Evaluation results on GLUE dev set. The teacher is fine-tuned DeBERTaV3-base model and the student is a DeBERTaV3-xsmall model.}
\vspace*{1mm}
\resizebox{0.7\textwidth}{!}{
\begin{tabular}{l|cccc}
\toprule
Method                                                    & RTE  & SST-2 & MRPC & STS-B \\ \midrule
LWD\textsubscript{xs}                                     & 80.2 & 93.9  & 90.2 & 91.0  \\
TED\textsubscript{xs} (Filters Learned on MNLI)           & 80.5 & 93.4  & 90.0 & 91.2  \\
TED\textsubscript{xs} (Filters Learned on QNLI)           & 81.6 & 93.5  & 89.1 & 91.4  \\
TED\textsubscript{xs} (Filters Learned on SST-2)          & 79.7 & \textbf{94.2}  & 90.2 & 90.8  \\
TED\textsubscript{xs} (Filters Learned on the Target Task) & \textbf{81.8} & \textbf{94.2}  & \textbf{90.4} & \textbf{91.3}  \\ \bottomrule
\end{tabular}}
\label{tb:task_relevancy}
\end{table*}

\begin{figure*}[htb!]
    \centering
    \vspace{0.1in}
    \includegraphics[width=0.85\linewidth]{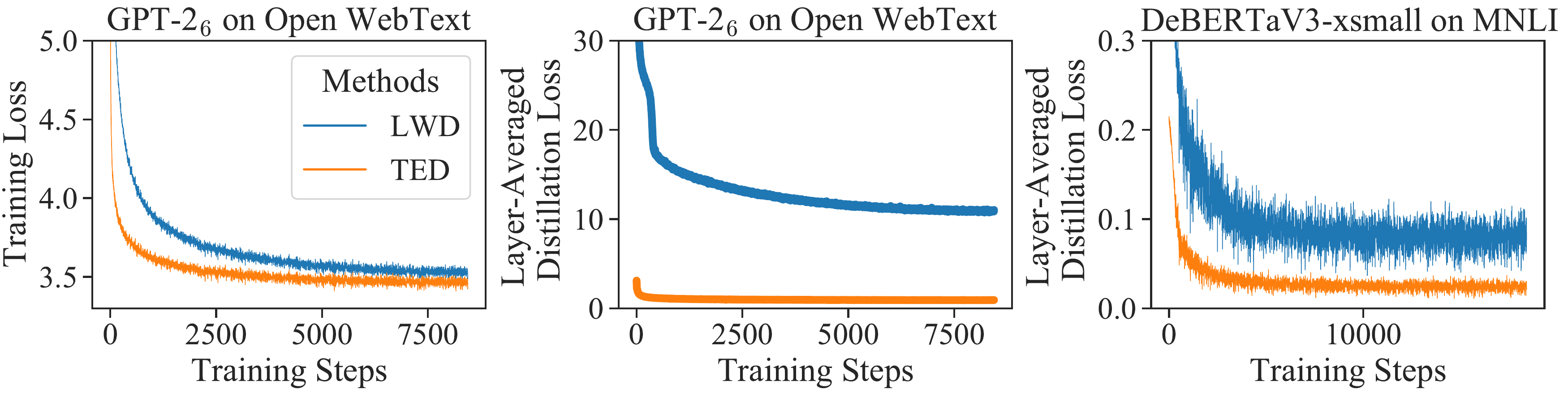}
    \caption{\textit{Left}: Training loss of the GPT-2$_6$ student on the target task (i.e., language modeling) on Open WebText. \textit{Middle and Right}: Distillation loss averaged by the number of layers of the GPT-2$_6$ student on Open WebText and the DeBERTaV3-xsmall student on MNLI, respectively.}
	\label{fig:all_losses}
\end{figure*}

\begin{table*}[htb!]
\centering
\caption{Evaluation results of different types of filter initialization. We evaluate the DeBERTaV3-xsmall student on the GLUE dev set and the GPT-2$_6$ student on the Open WebText test set. ``N/A'' is because we initialize $\Wcals$ of GPT-2${_6}$ from $\Wcalt$.}
\vspace*{1mm}
\resizebox{0.78\textwidth}{!}{
\begin{tabular}{l|cc|ccc|c}
\toprule
       &  $\Wcalt$       & $\Wcals$       & \multicolumn{4}{c}{Target Tasks}  \\\cmidrule{2-7}
Method &  Trained on  & Trained on  & MNLI-m/mm & SST-2 & RTE  & Open WebText  \\
       &  target task? & target task?  & Acc$\uparrow$       & Acc$\uparrow$   & Acc$\uparrow$   & ppl$\downarrow$         \\ \midrule
Abl.1  &  $\cmark$    & No $\Wcals$   & 87.4/86.9 & 92.0  & 77.2 & 31.82      \\ 
Abl.2  &  No $\Wcalt$    & $\xmark$   & 88.7/88.5 & 93.5  & 79.5 & 29.74       \\ 
Abl.3  &  $\cmark$    & $\xmark$   & 88.7/88.6 & 93.9  & 79.9 & 29.66      \\ 
Abl.4  &  No $\Wcalt$    & $\cmark$   & 88.8/88.6 & 94.0  & 81.5 & N/A         \\ 
{\ours}&  $\cmark$    & $\cmark$   & \textbf{88.8/88.7} & \textbf{94.2}  & \textbf{81.8} & \textbf{28.49}        \\ \bottomrule 
\end{tabular}}
\label{tb:ablation}
\end{table*}

Table~\ref{tb:task_relevancy} shows the evaluation results of a student model trained on the target task with their task-aware filters replaced by the filters trained on a different task. If the filters are trained on the target task, {\ours} shows consistent gains over LWD. In contrast, if the filters are trained on a different task, the gains become smaller and vary significantly across tasks, suggesting the task-aware filters can learn task-specific knowledge\footnote{The cause of high variance could be that the filters trained on tasks that are more similar to the target task perform better. For example, on the RTE task, filters trained on MNLI and QNLI perform better than those trained on SST-2, likely because the task-relevant knowledge can be transferred across NLI tasks.}. 

\subsection{{\ours} Alleviates Under-fitting and Eases Distillation}
\label{ana:main_claim}

Figure~\ref{fig:all_losses} (\textit{Left}) shows the training loss of the student on the target task (i.e., language modeling) during distillation. {\ours} leads to a faster convergence and a lower training loss than LWD, which suggests that {\ours} improves the fitting of the student on the target task. Figure~\ref{fig:all_losses} (\textit{Middle} and \textit{Right}) shows the distillation loss averaged by the number of layers. The distillation loss in {\ours} has a smaller magnitude and a lower variance than LWD. This suggests that {\ours} effectively eases the distillation.

\begin{table*}[htb!]
\centering
\caption{Evaluation results on GLUE dev set. The teacher is DeBERTaV3-large ($435$M), and the student is DeBERTaV3-xsmall ($70$M).}
\vspace{1mm}
\label{tb:capacity_gap}
\resizebox{0.53\textwidth}{!}{
\begin{tabular}{l|ccc}
\toprule
Method & MNLI-m/mm & SST-2 & RTE \\ \midrule
{\ours}\textsubscript{xs} (Teacher\textsubscript{base}) & \textbf{88.8}/88.7 & 94.2  & \textbf{81.8}  \\\midrule
Teacher\textsubscript{large} & 91.7/91.8 & 96.3  & 91.4  \\\midrule
Fine-tune\textsubscript{xs} (Teacher\textsubscript{large}) & 88.3/88.1 & 93.5  & 79.7 \\
KD\textsubscript{xs} (Teacher\textsubscript{large}) & 88.4/88.4 & 94.4  & 77.9 \\
LWD\textsubscript{xs} (Teacher\textsubscript{large}) & 88.5/88.5 & 93.6  & 79.4  \\ 
{\ours}\textsubscript{xs} (Teacher\textsubscript{large}) & \textbf{88.8/88.8} & \textbf{94.6}  & 81.4 \\ 
\bottomrule
\end{tabular}
}
\end{table*}

\begin{table*}[htb!]
\centering
\caption{Evaluation results of the GPT-2$_6$ student using different filter architectures.}
\vspace*{1mm}
\resizebox{0.8\textwidth}{!}{
\begin{tabular}{l|ccc|ccc}
\toprule
Filter Architectures & \multicolumn{3}{c|}{Zero-Shot} & \multicolumn{3}{c}{Transfer Learning} \\
& WikiText-103 & \multicolumn{2}{c|}{LAMBADA} & WikiText-103 & \multicolumn{2}{c}{LAMBADA} \\
& ppl$\downarrow$ & ppl$\downarrow$ & Acc$\uparrow$ & ppl$\downarrow$ & ppl$\downarrow$ & Acc$\uparrow$ \\ \midrule
Linear Projection & 48.12 & 87.27 & 22.99 & 19.03 & 48.63 & 32.08      \\
Two-layer MLP & \textbf{47.78} & \textbf{86.88} & \textbf{23.09} & 19.03 & 48.61 & 32.08 \\ 
One Subsequent Layer & 48.05 & 87.01 & 23.02 & \textbf{19.02} & 48.50 & 32.08 \\
All Subsequent Layers & 48.13 & 87.79 & 22.86 & 19.17 & \textbf{48.48} & \textbf{32.04} \\ \bottomrule
\end{tabular}}
\label{tb:complexity}
\end{table*}

\subsection{Contribution of the Filters}

To investigate the contribution of the filters, we initialize the filters with trained weights ($\cmark$), randomly initialized weights ($\xmark$), or no filters at all (No $\Wcalt$/$\Wcals$). Table~\ref{tb:ablation} shows that: 1) Using trained filters for the teacher significantly improves the distillation performance, as long as the student has a set of filters that can learn to match the teacher's filtered output. In other words, {\ours} can be still beneficial even if the student filters are not trained in Stage I but randomly initialized for Stage II. 2) If the student filters are initialized from trained weights instead of randomly initialized, the distillation performance can be further improved. 

\subsection{{\ours} Alleviates the Capacity Gap Issue}

Table~\ref{tb:capacity_gap} shows the performance of a DeBERTaV3-xsmall student ($70$M) distilled from a $6$ times larger DeBERTaV3-large model ($435$M), which contains $24$ layers and has a hidden dimension of $1024$. All implementation details are deferred to Appendix~\ref{app:capacity}. When using this large teacher, LWD performs worse, e.g., the student achieves only $79.4$ in RTE while achieving $80.2$ when using the DeBERTaV3-base teacher (Table~\ref{tb:deberta_glue}). In contrast, {\ours} maintains a comparable performance, e.g., the student achieves a $0.4$ points of gain on SST-2.

\subsection{Filter Architectures}
\label{ana:complexity}
Table~\ref{tb:complexity} shows the student performance when the filters are initialized with different architectures: 1) a linear projection; 2) a two-layer perceptron with GeLU as non-linearity \citep{hendrycks2016gaussian}; 3) layer(s) initialized from the weights of the subsequent layer(s), e.g., for the first layer of the model, we use the second (or from second to last) layer(s) as the filter. By introducing non-linearity in the filters, the zero-shot performance slightly improves while the transfer learning performance remains insensitive. Further increasing the filter complexity exhibits little benefits.

\subsection{Design of Distillation Loss}
\label{ana:feature_type}

If we keep the task specific head of each filter trained in Stage I and bring it to Stage II, then the filtered output would be a prediction probability distribution instead of a hidden representation. Then we can substitute the \texttt{MSE} between the two hidden representations in Eq.~\ref{eq:ted_hn_loss} with the KL-divergence between the two probability distributions. Table~\ref{tb:feature_type} shows that such an approach also shows noticeable improvements over the baselines. This suggests that the prediction probability can also preserve some task-specific knowledge. 

\begin{table}[htb!]
\centering
\caption{Evaluation results under different designs of the task-aware distillation loss. The teacher is DeBERTaV3-base and the student is DeBERTaV3-xsmall.}
\vspace{1.5mm}
\label{tb:feature_type}
\resizebox{0.35\textwidth}{!}{
\begin{tabular}{l|cc}
\toprule
Method & MNLI & SQuAD 2.0 \\
& m/mm & EM/F1 \\\midrule
KD & 88.5/88.1 & 81.0/84.2 \\
LWD & \textbf{88.8}/88.3 & 81.5/84.4 \\ \midrule
{\ours} (\texttt{KL})      & 88.6/88.7 & 81.9/84.7 \\
    {\ours} (\texttt{MSE})     & \textbf{88.8/88.7} & \textbf{82.0/84.9} \\\bottomrule
\end{tabular}}
\end{table}

\begin{figure*}[htb!]
    \centering
    \includegraphics[width=0.6\linewidth]{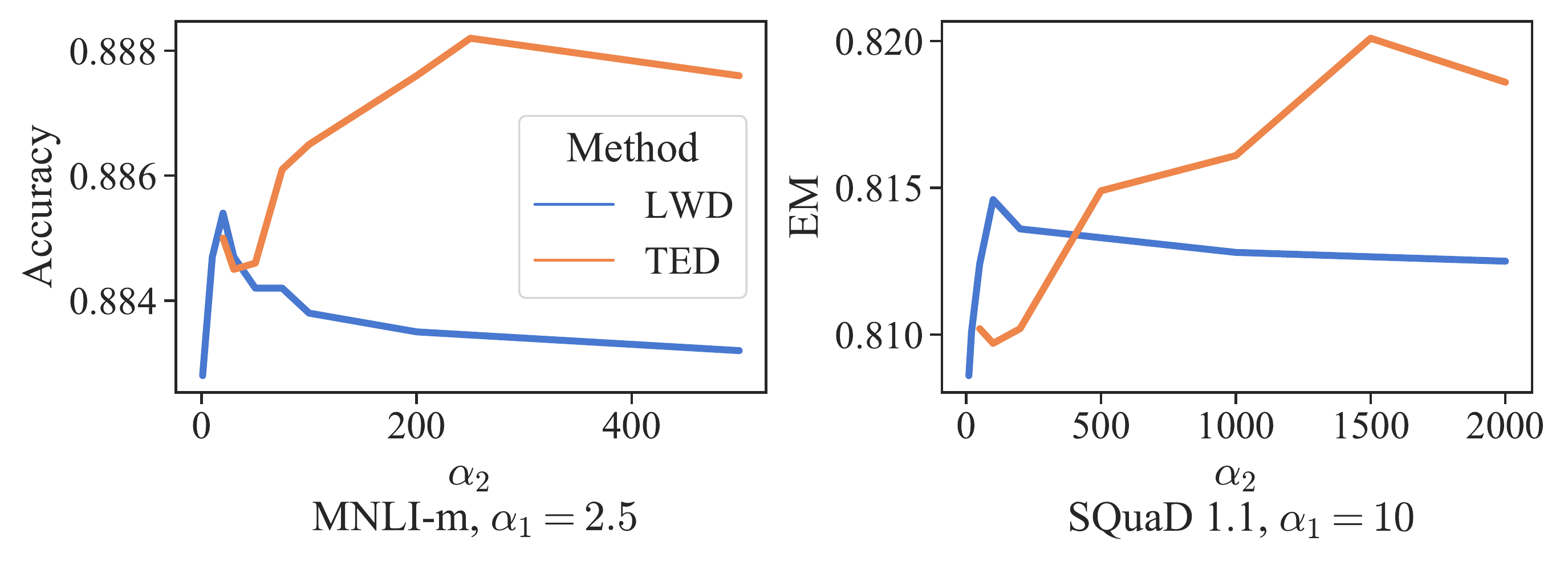}
        \caption{Evaluation performance of DeBERTaV3-xsmall under different values of $\alpha_2$.}
	\label{fig:alhpa}
\end{figure*}

\subsection{Hyper-parameter Study}

We further investigate whether {\ours} is sensitive to $\alpha_2$, the hyper-parameter that controls the strength of $\mathcal{D}_{\rm TED}$. Figure~\ref{fig:alhpa} shows the performance of the DeBERTaV3-xsmall student on MNLI-m and SQuAD v1.0 under different values of $\alpha_2$. {\ours} shows consistent gains over a wide range of values of $\alpha_2$.

%% file: 07-discussion.tex
\section{Discussion}
\label{sec:discussion}

\textbf{Computational Costs of {\ours}.} In the training phase, {\ours} incurs an additional computational overhead beyond what is required by layer-wise distillation (LWD). This is due to the training of task-aware filters in Stage I. However, this overhead is relatively moderate, accounting for approximately $10\%$ of the computational cost of LWD. This is because the number of filter parameters is around $2\%$-$4\%$ of the model parameters, and the training of the filters does not require any back-propagation on the model parameters. Despite the overhead during training, {\ours} retains the same inference speed as LWD during the model deployment phase. This is because all filters are discarded at this stage, with only model parameters being utilized for inference.

\textbf{Exploring Task-aware Distillation in Multi-task Setting.} We design {\ours} for \textit{task-specific distillation}, where a task-specific student is trained by distilling knowledge from a target task. However, task-specific distillation exhibits several practical limitations: 1) It lacks scalability as one needs to distill a new student for every new task. 2) Certain tasks have zero or few training samples, making them unsuitable for distillation. To resolve these limitations, one potential direction is to explore the idea of task-aware distillation in the \textit{multi-task setting} \citep{sanh2021multitask, longpre2023flan}. In this setting, one can leverage knowledge from hundreds of tasks to distill a single multi-task student which generalizes well on various seen and unseen tasks. One possible strategy, for example, is to design multiple filters, each serving as an expert specialized in extracting knowledge from a group of relevant tasks. During the distillation process, each input sample can be routed to its most task-relevant filter(s). 

\textbf{Filtering and Distilling Knowledge from Large Language Models.} Pre-trained large language models (LLMs), with up to hundreds of billions of parameters, have demonstrated remarkable generalizability on a wide range of tasks. How to effectively transfer the knowledge from these powerful LLMs into smaller models has therefore become an area of research interest \citep{hsieh2023distilling,jiang2023lion,alpaca,peng2023instruction}. However, it is challenging to directly apply TED, an layer-wise distillation (LWD) approach, to LLMs due to two primary reasons: 1) LWD requires the access to the layer-wise hidden representations, making it incompatible with models that are closed-source. 2) The computational cost of LWD scales with the model depth and hidden dimension, which is prohibitively expensive in such models. Yet, the underlying idea of selecting and transferring task-relevant knowledge could be useful for LLM distillation, particularly given the significant teacher-student capacity gap. For example, a possible strategy is to use a LLM teacher to generate task-relevant input and output samples in a controllable manner, and then use these samples to distill the student model.

%% file: 08-conclusion.tex
\section{Conclusion}
\label{sec:conclusion}
 
Layer-wise distillation is challenging as the student may struggle to mimic the hidden representations of a much larger teacher. We propose {\ours}, which first learns task-aware filters to extract the task-specific knowledge from the teacher and the student, then minimizes the discrepancy between the filtered outputs. This encourages the student to learn filtered knowledge, which contains more task-relevant signals. Our experiments verify that the filters can effectively capture task-specific knowledge and ease layer-wise distillation. 

%% file: appendix.tex
\section{Appendix}

\subsection{Language Modeling Experiment}
\subsubsection{Data}
\label{app:gpt_data}
Open WebText is an open source effort to reproduce OpenAI’s WebText dataset. The dataset is created by extracting Reddit post urls from the Reddit submissions dataset\footnote{https://files.pushshift.io/reddit/submissions/}. These links are then deduplicated, filtered to exclude non-html content, and shuffled randomly. Near-duplicate documents are identified using local-sensitivity hashing. They are hashed into sets of $5$-grams and all documents that had a similarity threshold of greater than $0.5$ were removed. All language modeling datasets were tokenized based on byte-level BPE \citep{sennrich2015neural} with a vocabulary size of $50257$ \citep{radford2019language}. The max sequence length of the input training sample is $1024$. 

\subsubsection{Training}
\label{app:gpt_filter}
Our implementation is based on Huggingface Transformers \footnote{https://github.com/huggingface/transformers/tree/v4.17.0}. The GPT-2 base model consists $12$ layers and has $12$ attention heads in each attention module. The input and intermediate hidden dimension in the feed-forward network is $768$ and $1024$, respectively. We use mixed precision training and train on $8$ $80$G Nvidia A100 GPUs. Detailed hyper-parameters are summarized in Table~\ref{tb:gpt_params}. 

\begin{table}[htb!]
\centering
\caption{Hyper-parameters for training GPT-2$_6$ on Open WebText.}
\begin{tabular}{l|cc}
\toprule
Hyper-parameters    & Stage I & Stage II \\ \midrule
Dropout             &      0.1           & 0.1      \\
Warmup Ratio        &      0.05 & 0.05    \\
Learning Rates      &      0.00025 & 0.00025  \\
Batch Size          &      4000 & 4000     \\
Weight Decay        &      0 & 0   \\
Training Epochs     &      1             &       4      \\
Learning Rate Decay &      Linear & Linear         \\
Adam $\epsilon$     &      $1\times10^{-6}$ &  $1\times10^{-6}$   \\
Adam $\beta_1$      &      0.9 & 0.9   \\
Adam $\beta_2$      &      0.98 & 0.98       \\ \bottomrule
\end{tabular}
\label{tb:gpt_params}
\end{table}

\subsection{Natural Language Understanding Experiment}
\subsubsection{Data}
\label{app:nlu_data}

GLUE is a commonly used natural language understanding benchmark containing nine tasks. The benchmark includes question answering \citep{squad1}, linguistic acceptability (CoLA, \citealt{cola2018}), sentiment analysis (SST, \citealt{sst2013}), text similarity (STS-B, \citealt{sts-b2017}), paraphrase detection (MRPC, \citealt{mrpc2005}), and natural language inference (RTE \& MNLI, \citealt{rte1,rte2,rte3,rte5,mnli2018}) tasks. Details of the GLUE benchmark, including tasks, statistics, and evaluation metrics, are summarized in Table~\ref{tab:glue}. 

SQuAD 1.1/2.0 is the Stanford Question Answering Dataset (SQuAD) v1.1 and v2.0 \citep{rajpurkar2018know, rajpurkar2016squad}, two popular machine reading comprehension benchmarks from approximately $500$ Wikipedia articles with questions and answers obtained by crowdsourcing. The SQuAD v2.0 dataset includes unanswerable questions about the same paragraphs.

\subsubsection{Model}
\label{app:nlu_model}

We initialize the teacher for each target task as a DeBERTaV3-base model fine-tuned on the target task. We fine-tune the model by adding a target task classification head on top of the last layer. The detailed hyper-parameters are listed in Table~\ref{tb:deberta_params}. We initialize the student for each target task as a pre-trained DeBERTaV3-xsmall model.

\subsubsection{Training}
\label{app:nlu_training}

We follow the hyper-parameter configurations listed in Table~\ref{tb:deberta_params} for both the Stage I and Stage II training. Our implementation is based on Huggingface Transformers. We use mixed precision training and train on $8$ $32$G Nvidia V100 GPUs. 

For Stage I, we empirically observe that if we first fine-tune the student on the target task, then train the filters on top of the fine-tuned student, the distillation performance would improve. We hypothesize that the student filters can learn to capture more task-relevant knowledge if the student is properly initialized on the target task. As a result, we also fine-tune the student model following the hyper-parameter configuration listed in Table~\ref{tb:deberta_params} before Stage I. As shown in Table~\ref{tb:student_init_ablation}, initializing the student model with fine-tuned weights will not largely influence the final distillation performance.

\begin{table*}[htb!]
\centering
\caption{Performance comparison of initializing the student with fine-tuned and pre-trained weights.}
\resizebox{0.65\textwidth}{!}{
\begin{tabular}{l|c|cccccc}
\toprule
Method &  $\Theta_s$    & MNLI-m/mm & QQP & QNLI & SST-2 & RTE & Avg\\
      &  Fine-tuned?   & Acc       & Acc & Acc  &Acc    & Acc & Score \\ \midrule
LWD    &  $\xmark$      & 88.8/88.3 & 91.8& 92.9 &93.9   & 80.2 & 89.5\\
Abl.   &  $\cmark$      & 88.7/88.5 & 92.0& 92.8 &93.5   & 79.5 & 89.3 \\ \bottomrule
\end{tabular}}
\label{tb:student_init_ablation}
\end{table*}

\subsubsection{BERT Experiments}
\label{app:nlu_bert}

\textbf{Model.} We initialize the teacher model with a pre-trained $12$-layer BERT-base model that has been fine-tuned on the target task (BERT-base$_{12}$). The teacher model contains $110$M parameters and has a hidden dimension of $d_t=768$. We initialize the student model with $6$ selected layers from the fine-tuned teacher model (BERT-base$_6$). Specifically, we define the layer mapping function $M(k) = 2k-1$ for $k \leq K/2$ and $M(k) = 2k$ for $k > K/2$, which is the same as \citealt{sanh2019distilbert}. The fine-tuning hyper-parameters are listed in Table~\ref{tb:bert_params}.

\textbf{Stage I.} We initialize each task-aware filter of the teacher with size $d_t\times d_t$. We fix the fine-tuned teacher and train the filters following the hyper-parameter configurations listed in Table~\ref{tb:bert_params}. We directly take the trained $k$-th filter of the teacher as the $k$-th filter of the student without further training. 

\textbf{Stage II.} We distill the student model and its filters following the hyper-parameter configurations listed in Table~\ref{tb:bert_params}. Our implementation is based on Huggingface Transformers. We conduct all experiments using mixed precision training on $8$ $32$G Nvidia V100 GPUs. 

\begin{table}[htb!]
\centering
\caption{Hyper-parameters for fine-tuning BERT-base$_{12}$ on MNLI.}
\begin{tabular}{l|c}
\toprule
Hyper-parameters      & BERT-base \\ \midrule
Dropout of Task Layer & 0.1        \\
Warmup Steps          & 1000        \\
Learning Rates        & $3\times 10^{-5}$        \\
Batch Size            & 32        \\
Weight Decay          & 0       \\
Training Epochs       & 3    \\
Learning Rate Decay   & Linear   \\
Adam $\epsilon$       & $1\times10^{-6}$ \\
Adam $\beta_1$        & 0.9          \\
Adam $\beta_2$        & 0.98      \\ \bottomrule
\end{tabular}
\label{tb:bert_params}
\end{table}

\subsection{Experiments in Analysis}
\label{app:ana}

\subsubsection{Experiments in Section~\ref{ana:main_claim}}
\label{app:capacity}

\textbf{Model.} We initialize the teacher model with a pre-trained $24$-layer DeBERTaV3-large model that has been fine-tuned on the target task. The teacher model contains $435$M parameters and has a hidden dimension $d_t=1024$. We initialize the student model with a $12$-layer DeBERTav3-xsmall model. The student model contains $70$M parameters and has a hidden dimension $d_s=384$. We define the layer mapping function $M(k) = 2k-1$ for $k \leq K/2$ and $M(k) = 2k$ for $k > K/2$, which is the same as \citealt{sanh2019distilbert}. The fine-tuning hyper-parameters are listed in Table~\ref{tb:deberta_params}.

\textbf{Stage I.} We initialize each filter of the teacher model with the size $d_t\times d_t$ and each filter of the student model with the size $d_s \times d_t$. We fix the model parameters of the teacher and the student and train their filters following the hyper-parameters summarized in Table~\ref{tb:deberta_params}.

\textbf{Stage II.} We distill the student model and its filters following the hyper-parameters listed in Table~\ref{tb:deberta_params}. Our implementation is based on Huggingface Transformers. We conduct all experiments using mixed precision training on $8$ $32$G Nvidia V100 GPUs. 

\begin{table*}[htb!]
\centering
\small
\caption{Hyper-parameters for fine-tuning DeBERTaV3 models on the downstream tasks.}
\begin{tabular}{l|ccc}
\toprule
Hyper-parameters      & DeBERTaV3-large & DeBERTaV3-base & DeBERTaV3-xsmall \\ \midrule
Dropout of Task Layer & $\{0.05, 0.1\}$                & $\{0.05, 0.1, 0.15\}$     &  $\{0.05, 0.1, 0.15\}$    \\
Learning Rates        &  $\{6,7,10\}\times 10^{-6}$               & $\{1,1.5,2,2.5,3,4,5\}\times 10^{-5}$               & $\{3,3.5,5,6,8,9\}\times 10^{-5}$   \\
Batch Size            &  \{32, 64\}     & \{12, 16, 32, 64\} & \{12,16,32, 64\}                 \\
Weight Decay          &  0            &  0            &  0             \\
Training Epochs       & \{2,6,8\}       & \{2,3,6,8\}    & \{2,3,6,8\}          \\
Learning Rate Decay   & Linear          & Linear         &   Linear      \\
Adam $\epsilon$       & $1\times10^{-6}$&$1\times10^{-6}$& $1\times10^{-6}$ \\
Adam $\beta_1$        & 0.9             & 0.9            & 0.9              \\
Adam $\beta_2$        & 0.98            & 0.98           & 0.98  \\
\bottomrule
\end{tabular}
\label{tb:deberta_params}
\end{table*}

\begin{table*}[htb]
    \caption{Summary of the GLUE benchmark.}
	\begin{center}
		\begin{tabular}{l|l|c|c|c|c|c}
			\toprule 
			\bf Corpus &Task& \#Train & \#Dev & \#Test   & \#Label &Metrics\\ \midrule
			\multicolumn{6}{@{\hskip1pt}r@{\hskip1pt}}{Single-Sentence Classification (GLUE)} \\ \hline
			CoLA & Acceptability&8.5k & 1k & 1k & 2 & Matthews corr\\ \hline
			SST & Sentiment&67k & 872 & 1.8k & 2 & Accuracy\\ \midrule
			\multicolumn{6}{@{\hskip1pt}r@{\hskip1pt}}{Pairwise Text Classification (GLUE)} \\ \hline
			MNLI & NLI& 393k& 20k & 20k& 3 & Accuracy\\ \hline
            RTE & NLI &2.5k & 276 & 3k & 2 & Accuracy \\ \hline
			QQP & Paraphrase&364k & 40k & 391k& 2 & Accuracy/F1\\ \hline
            MRPC & Paraphrase &3.7k & 408 & 1.7k& 2&Accuracy/F1\\ \hline
			QNLI & QA/NLI& 108k &5.7k&5.7k&2& Accuracy\\ \midrule
			\multicolumn{6}{@{\hskip1pt}r@{\hskip1pt}}{Text Similarity (GLUE)} \\ \hline
			STS-B & Similarity &7k &1.5k& 1.4k &1 & Pearson/Spearman corr\\ \bottomrule
		\end{tabular}
	\end{center}
	\label{tab:glue}
\end{table*}

\subsection{Discussion on the Model Initialization}
\label{app:initialization}

The model initialization is critical to the learning of the task-aware filters. If the model parameters have not been properly initialized and the filters are directly trained upon such parameters, the filters may fail to learn sufficient task-relevant knowledge and become useless. Below we list our recommended practices for model initialization under different scenarios:

\noindent \textbf{Distillation in the pre-training setting.} This setting considers a pre-trained model as the teacher and produces a pre-trained model as the student.

\textit{Case 1}. If there exists a pre-trained model with the desired student architecture, we can directly initialize the student with its weights and proceed to Stage I.

\textit{Case 2}. If there does not exist a pre-trained model with the desired student architecture, we consider the following three cases: \textit{Case 2.1}. If there is sufficient computational budget, we can pre-train the student from scratch and then proceed to Stage I. \textit{Case 2.2}. If there is no pre-training budget but the desired student architecture is a shallow version of the teacher (like in the GPT-2$_6$ case), we can initialize the student with a subset of teacher layers. We can directly adopt the filters of the teacher at the corresponding layers as the filters of the student, and proceed to Stage II. \textit{Case 2.3}. Otherwise, we recommend directly proceeding to Stage II.

\noindent \textbf{Distillation in the fine-tuning setting.} This setting considers a fine-tuned model as the teacher and produces a fine-tuned model as the student.

\textit{Case 1}. If there exists a pre-trained model with the desired student architecture (like the DeBERTaV3-xsmall case), we can fine-tune the pre-trained model on the target task, initialize the student with its weights, and proceed to Stage I.

\textit{Case 2}. If there does not exist a pre-trained model with the desired student architecture, we consider the following three cases: \textit{Case 2.1}. If there is sufficient computational budget, we can pre-train and fine-tune the student from scratch and then proceed to Stage I. \textit{Case 2.2}. If there is limited computational budget, but the desired student architecture is a shallow version of the teacher (like the BERT-base$_6$ case), we can initialize the student with a subset of teacher layers. We directly adopt the filters of the teacher at the corresponding layers as the filters of the student, and proceed to Stage II. \textit{Case 2.3}. Otherwise, we recommend directly proceeding to Stage II. 

%% file: icml2023.bbl
\begin{thebibliography}{53}
\providecommand{\natexlab}[1]{#1}
\providecommand{\url}[1]{\texttt{#1}}
\expandafter\ifx\csname urlstyle\endcsname\relax
  \providecommand{\doi}[1]{doi: #1}\else
  \providecommand{\doi}{doi: \begingroup \urlstyle{rm}\Url}\fi

\bibitem[Bar-Haim et~al.(2006)Bar-Haim, Dagan, Dolan, Ferro, and
  Giampiccolo]{rte2}
Bar-Haim, R., Dagan, I., Dolan, B., Ferro, L., and Giampiccolo, D.
\newblock The second {PASCAL} recognising textual entailment challenge.
\newblock In \emph{Proceedings of the Second {PASCAL} Challenges Workshop on
  Recognising Textual Entailment}, 2006.

\bibitem[Bentivogli et~al.(2009)Bentivogli, Dagan, Dang, Giampiccolo, and
  Magnini]{rte5}
Bentivogli, L., Dagan, I., Dang, H.~T., Giampiccolo, D., and Magnini, B.
\newblock The fifth pascal recognizing textual entailment challenge.
\newblock In \emph{In Proc Text Analysis Conference (TAC’09)}, 2009.

\bibitem[Brown et~al.(2020)Brown, Mann, Ryder, Subbiah, Kaplan, Dhariwal,
  Neelakantan, Shyam, Sastry, Askell, Agarwal, Herbert{-}Voss, Krueger,
  Henighan, Child, Ramesh, Ziegler, Wu, Winter, Hesse, Chen, Sigler, Litwin,
  Gray, Chess, Clark, Berner, McCandlish, Radford, Sutskever, and
  Amodei]{brown2020language}
Brown, T.~B., Mann, B., Ryder, N., Subbiah, M., Kaplan, J., Dhariwal, P.,
  Neelakantan, A., Shyam, P., Sastry, G., Askell, A., Agarwal, S.,
  Herbert{-}Voss, A., Krueger, G., Henighan, T., Child, R., Ramesh, A.,
  Ziegler, D.~M., Wu, J., Winter, C., Hesse, C., Chen, M., Sigler, E., Litwin,
  M., Gray, S., Chess, B., Clark, J., Berner, C., McCandlish, S., Radford, A.,
  Sutskever, I., and Amodei, D.
\newblock Language models are few-shot learners.
\newblock In Larochelle, H., Ranzato, M., Hadsell, R., Balcan, M., and Lin, H.
  (eds.), \emph{Advances in Neural Information Processing Systems 33: Annual
  Conference on Neural Information Processing Systems 2020, NeurIPS 2020,
  December 6-12, 2020, virtual}, 2020.

\bibitem[Cer et~al.(2017)Cer, Diab, Agirre, Lopez-Gazpio, and
  Specia]{sts-b2017}
Cer, D., Diab, M., Agirre, E., Lopez-Gazpio, I., and Specia, L.
\newblock {S}em{E}val-2017 task 1: Semantic textual similarity multilingual and
  crosslingual focused evaluation.
\newblock In \emph{Proceedings of the 11th International Workshop on Semantic
  Evaluation ({S}em{E}val-2017)}, pp.\  1--14, Vancouver, Canada, 2017.
  Association for Computational Linguistics.
\newblock \doi{10.18653/v1/S17-2001}.

\bibitem[Clark et~al.(2020)Clark, Luong, Le, and Manning]{clark2020electra}
Clark, K., Luong, M., Le, Q.~V., and Manning, C.~D.
\newblock {ELECTRA:} pre-training text encoders as discriminators rather than
  generators.
\newblock In \emph{8th International Conference on Learning Representations,
  {ICLR} 2020, Addis Ababa, Ethiopia, April 26-30, 2020}. OpenReview.net, 2020.

\bibitem[Dagan et~al.(2006)Dagan, Glickman, and Magnini]{rte1}
Dagan, I., Glickman, O., and Magnini, B.
\newblock The pascal recognising textual entailment challenge.
\newblock In \emph{Proceedings of the First International Conference on Machine
  Learning Challenges: Evaluating Predictive Uncertainty Visual Object
  Classification, and Recognizing Textual Entailment}, MLCW'05, pp.\  177--190,
  Berlin, Heidelberg, 2006. Springer-Verlag.
\newblock ISBN 3-540-33427-0, 978-3-540-33427-9.
\newblock \doi{10.1007/11736790_9}.

\bibitem[Dalvi et~al.(2020)Dalvi, Sajjad, Durrani, and
  Belinkov]{dalvi2020analyzing}
Dalvi, F., Sajjad, H., Durrani, N., and Belinkov, Y.
\newblock Analyzing redundancy in pretrained transformer models.
\newblock In \emph{Proceedings of the 2020 Conference on Empirical Methods in
  Natural Language Processing (EMNLP)}, pp.\  4908--4926, Online, 2020.
  Association for Computational Linguistics.
\newblock \doi{10.18653/v1/2020.emnlp-main.398}.

\bibitem[Devlin et~al.(2019)Devlin, Chang, Lee, and Toutanova]{devlin2018bert}
Devlin, J., Chang, M.-W., Lee, K., and Toutanova, K.
\newblock {BERT}: Pre-training of deep bidirectional transformers for language
  understanding.
\newblock In \emph{Proceedings of the 2019 Conference of the North {A}merican
  Chapter of the Association for Computational Linguistics: Human Language
  Technologies, Volume 1 (Long and Short Papers)}, pp.\  4171--4186,
  Minneapolis, Minnesota, 2019. Association for Computational Linguistics.
\newblock \doi{10.18653/v1/N19-1423}.

\bibitem[Dolan \& Brockett(2005)Dolan and Brockett]{mrpc2005}
Dolan, W.~B. and Brockett, C.
\newblock Automatically constructing a corpus of sentential paraphrases.
\newblock In \emph{Proceedings of the Third International Workshop on
  Paraphrasing ({IWP}2005)}, 2005.

\bibitem[Durrani et~al.(2020)Durrani, Sajjad, Dalvi, and
  Belinkov]{durrani2020analyzing}
Durrani, N., Sajjad, H., Dalvi, F., and Belinkov, Y.
\newblock Analyzing individual neurons in pre-trained language models.
\newblock In \emph{Proceedings of the 2020 Conference on Empirical Methods in
  Natural Language Processing (EMNLP)}, pp.\  4865--4880, Online, 2020.
  Association for Computational Linguistics.
\newblock \doi{10.18653/v1/2020.emnlp-main.395}.

\bibitem[Giampiccolo et~al.(2007)Giampiccolo, Magnini, Dagan, and Dolan]{rte3}
Giampiccolo, D., Magnini, B., Dagan, I., and Dolan, B.
\newblock The third {PASCAL} recognizing textual entailment challenge.
\newblock In \emph{Proceedings of the {ACL}-{PASCAL} Workshop on Textual
  Entailment and Paraphrasing}, pp.\  1--9, Prague, 2007. Association for
  Computational Linguistics.

\bibitem[Gokaslan et~al.(2019)Gokaslan, Cohen, Pavlick, and
  Tellex]{gokaslan2019openwebtext}
Gokaslan, A., Cohen, V., Pavlick, E., and Tellex, S.
\newblock Openwebtext corpus, 2019.

\bibitem[He et~al.(2020)He, Liu, Gao, and Chen]{he2020deberta}
He, P., Liu, X., Gao, J., and Chen, W.
\newblock Deberta: Decoding-enhanced bert with disentangled attention.
\newblock \emph{arXiv preprint arXiv:2006.03654}, 2020.

\bibitem[He et~al.(2023)He, Gao, and Chen]{he2023debertav}
He, P., Gao, J., and Chen, W.
\newblock De{BERT}av3: Improving de{BERT}a using {ELECTRA}-style pre-training
  with gradient-disentangled embedding sharing.
\newblock In \emph{The Eleventh International Conference on Learning
  Representations}, 2023.
\newblock URL \url{https://openreview.net/forum?id=sE7-XhLxHA}.

\bibitem[Hendrycks \& Gimpel(2016)Hendrycks and Gimpel]{hendrycks2016gaussian}
Hendrycks, D. and Gimpel, K.
\newblock Gaussian error linear units (gelus).
\newblock \emph{arXiv preprint arXiv:1606.08415}, 2016.

\bibitem[Hinton et~al.(2015)Hinton, Vinyals, and Dean]{hinton2015distilling}
Hinton, G., Vinyals, O., and Dean, J.
\newblock Distilling the knowledge in a neural network.
\newblock \emph{arXiv preprint arXiv:1503.02531}, 2015.

\bibitem[Hou et~al.(2020)Hou, Huang, Shang, Jiang, Chen, and
  Liu]{hou2020dynabert}
Hou, L., Huang, Z., Shang, L., Jiang, X., Chen, X., and Liu, Q.
\newblock Dynabert: Dynamic {BERT} with adaptive width and depth.
\newblock In Larochelle, H., Ranzato, M., Hadsell, R., Balcan, M., and Lin, H.
  (eds.), \emph{Advances in Neural Information Processing Systems 33: Annual
  Conference on Neural Information Processing Systems 2020, NeurIPS 2020,
  December 6-12, 2020, virtual}, 2020.

\bibitem[Hsieh et~al.(2023)Hsieh, Li, Yeh, Nakhost, Fujii, Ratner, Krishna,
  Lee, and Pfister]{hsieh2023distilling}
Hsieh, C.-Y., Li, C.-L., Yeh, C.-K., Nakhost, H., Fujii, Y., Ratner, A.,
  Krishna, R., Lee, C.-Y., and Pfister, T.
\newblock Distilling step-by-step! outperforming larger language models with
  less training data and smaller model sizes.
\newblock \emph{arXiv preprint arXiv:2305.02301}, 2023.

\bibitem[Jiang et~al.(2023)Jiang, Chan, Chen, and Wang]{jiang2023lion}
Jiang, Y., Chan, C., Chen, M., and Wang, W.
\newblock Lion: Adversarial distillation of closed-source large language model.
\newblock \emph{arXiv preprint arXiv:2305.12870}, 2023.

\bibitem[Jiao et~al.(2020)Jiao, Yin, Shang, Jiang, Chen, Li, Wang, and
  Liu]{jiao2019tinybert}
Jiao, X., Yin, Y., Shang, L., Jiang, X., Chen, X., Li, L., Wang, F., and Liu,
  Q.
\newblock {T}iny{BERT}: Distilling {BERT} for natural language understanding.
\newblock In \emph{Findings of the Association for Computational Linguistics:
  EMNLP 2020}, pp.\  4163--4174, Online, 2020. Association for Computational
  Linguistics.
\newblock \doi{10.18653/v1/2020.findings-emnlp.372}.

\bibitem[Liang et~al.(2023)Liang, Jiang, Li, Tang, Yin, and
  Zhao]{liang2023homodistil}
Liang, C., Jiang, H., Li, Z., Tang, X., Yin, B., and Zhao, T.
\newblock Homodistil: Homotopic task-agnostic distillation of pre-trained
  transformers.
\newblock \emph{arXiv preprint arXiv:2302.09632}, 2023.

\bibitem[Liang et~al.(2020)Liang, Hao, Shen, Zhou, Chen, Chen, and
  Carin]{liang2020mixkd}
Liang, K.~J., Hao, W., Shen, D., Zhou, Y., Chen, W., Chen, C., and Carin, L.
\newblock Mixkd: Towards efficient distillation of large-scale language models.
\newblock \emph{arXiv preprint arXiv:2011.00593}, 2020.

\bibitem[Longpre et~al.(2023)Longpre, Hou, Vu, Webson, Chung, Tay, Zhou, Le,
  Zoph, Wei, et~al.]{longpre2023flan}
Longpre, S., Hou, L., Vu, T., Webson, A., Chung, H.~W., Tay, Y., Zhou, D., Le,
  Q.~V., Zoph, B., Wei, J., et~al.
\newblock The flan collection: Designing data and methods for effective
  instruction tuning.
\newblock \emph{arXiv preprint arXiv:2301.13688}, 2023.

\bibitem[Loshchilov \& Hutter(2019)Loshchilov and
  Hutter]{loshchilov2017decoupled}
Loshchilov, I. and Hutter, F.
\newblock Decoupled weight decay regularization.
\newblock In \emph{7th International Conference on Learning Representations,
  {ICLR} 2019, New Orleans, LA, USA, May 6-9, 2019}. OpenReview.net, 2019.

\bibitem[Merity et~al.(2017)Merity, Xiong, Bradbury, and
  Socher]{merity2016pointer}
Merity, S., Xiong, C., Bradbury, J., and Socher, R.
\newblock Pointer sentinel mixture models.
\newblock In \emph{5th International Conference on Learning Representations,
  {ICLR} 2017, Toulon, France, April 24-26, 2017, Conference Track
  Proceedings}. OpenReview.net, 2017.

\bibitem[Nagel(2016)]{nagel2016cc}
Nagel, S.
\newblock Cc-news.
\newblock \emph{URL: http://web. archive. org/save/http://commoncrawl.
  org/2016/10/newsdatasetavailable}, 2016.

\bibitem[Paperno et~al.(2016)Paperno, Kruszewski, Lazaridou, Pham, Bernardi,
  Pezzelle, Baroni, Boleda, and Fern{\'a}ndez]{paperno2016lambada}
Paperno, D., Kruszewski, G., Lazaridou, A., Pham, N.~Q., Bernardi, R.,
  Pezzelle, S., Baroni, M., Boleda, G., and Fern{\'a}ndez, R.
\newblock The {LAMBADA} dataset: Word prediction requiring a broad discourse
  context.
\newblock In \emph{Proceedings of the 54th Annual Meeting of the Association
  for Computational Linguistics (Volume 1: Long Papers)}, pp.\  1525--1534,
  Berlin, Germany, 2016. Association for Computational Linguistics.
\newblock \doi{10.18653/v1/P16-1144}.

\bibitem[Peng et~al.(2023)Peng, Li, He, Galley, and Gao]{peng2023instruction}
Peng, B., Li, C., He, P., Galley, M., and Gao, J.
\newblock Instruction tuning with gpt-4.
\newblock \emph{arXiv preprint arXiv:2304.03277}, 2023.

\bibitem[Radford et~al.(2019)Radford, Wu, Child, Luan, Amodei, Sutskever,
  et~al.]{radford2019language}
Radford, A., Wu, J., Child, R., Luan, D., Amodei, D., Sutskever, I., et~al.
\newblock Language models are unsupervised multitask learners.
\newblock \emph{OpenAI blog}, 1\penalty0 (8):\penalty0 9, 2019.

\bibitem[Raffel et~al.(2019)Raffel, Shazeer, Roberts, Lee, Narang, Matena,
  Zhou, Li, and Liu]{raffel2019exploring}
Raffel, C., Shazeer, N., Roberts, A., Lee, K., Narang, S., Matena, M., Zhou,
  Y., Li, W., and Liu, P.~J.
\newblock Exploring the limits of transfer learning with a unified text-to-text
  transformer.
\newblock \emph{arXiv preprint arXiv:1910.10683}, 2019.

\bibitem[Rajpurkar et~al.(2016{\natexlab{a}})Rajpurkar, Zhang, Lopyrev, and
  Liang]{rajpurkar2016squad}
Rajpurkar, P., Zhang, J., Lopyrev, K., and Liang, P.
\newblock {SQ}u{AD}: 100,000+ questions for machine comprehension of text.
\newblock In \emph{Proceedings of the 2016 Conference on Empirical Methods in
  Natural Language Processing}, pp.\  2383--2392, Austin, Texas,
  2016{\natexlab{a}}. Association for Computational Linguistics.
\newblock \doi{10.18653/v1/D16-1264}.

\bibitem[Rajpurkar et~al.(2016{\natexlab{b}})Rajpurkar, Zhang, Lopyrev, and
  Liang]{squad1}
Rajpurkar, P., Zhang, J., Lopyrev, K., and Liang, P.
\newblock {SQ}u{AD}: 100,000+ questions for machine comprehension of text.
\newblock In \emph{Proceedings of the 2016 Conference on Empirical Methods in
  Natural Language Processing}, pp.\  2383--2392, Austin, Texas,
  2016{\natexlab{b}}. Association for Computational Linguistics.
\newblock \doi{10.18653/v1/D16-1264}.

\bibitem[Rajpurkar et~al.(2018)Rajpurkar, Jia, and Liang]{rajpurkar2018know}
Rajpurkar, P., Jia, R., and Liang, P.
\newblock Know what you don{'}t know: Unanswerable questions for {SQ}u{AD}.
\newblock In \emph{Proceedings of the 56th Annual Meeting of the Association
  for Computational Linguistics (Volume 2: Short Papers)}, pp.\  784--789,
  Melbourne, Australia, 2018. Association for Computational Linguistics.
\newblock \doi{10.18653/v1/P18-2124}.

\bibitem[Romero et~al.(2015)Romero, Ballas, Kahou, Chassang, Gatta, and
  Bengio]{romero2014fitnets}
Romero, A., Ballas, N., Kahou, S.~E., Chassang, A., Gatta, C., and Bengio, Y.
\newblock Fitnets: Hints for thin deep nets.
\newblock In Bengio, Y. and LeCun, Y. (eds.), \emph{3rd International
  Conference on Learning Representations, {ICLR} 2015, San Diego, CA, USA, May
  7-9, 2015, Conference Track Proceedings}, 2015.

\bibitem[Sanh et~al.(2019)Sanh, Debut, Chaumond, and Wolf]{sanh2019distilbert}
Sanh, V., Debut, L., Chaumond, J., and Wolf, T.
\newblock Distilbert, a distilled version of bert: smaller, faster, cheaper and
  lighter.
\newblock \emph{arXiv preprint arXiv:1910.01108}, 2019.

\bibitem[Sanh et~al.(2021)Sanh, Webson, Raffel, Bach, Sutawika, Alyafeai,
  Chaffin, Stiegler, Scao, Raja, et~al.]{sanh2021multitask}
Sanh, V., Webson, A., Raffel, C., Bach, S.~H., Sutawika, L., Alyafeai, Z.,
  Chaffin, A., Stiegler, A., Scao, T.~L., Raja, A., et~al.
\newblock Multitask prompted training enables zero-shot task generalization.
\newblock \emph{arXiv preprint arXiv:2110.08207}, 2021.

\bibitem[Sennrich et~al.(2016)Sennrich, Haddow, and Birch]{sennrich2015neural}
Sennrich, R., Haddow, B., and Birch, A.
\newblock Neural machine translation of rare words with subword units.
\newblock In \emph{Proceedings of the 54th Annual Meeting of the Association
  for Computational Linguistics (Volume 1: Long Papers)}, pp.\  1715--1725,
  Berlin, Germany, 2016. Association for Computational Linguistics.
\newblock \doi{10.18653/v1/P16-1162}.

\bibitem[Shi et~al.(2021)Shi, Song, Zhou, Li, and Li]{shi2021follow}
Shi, W., Song, Y., Zhou, H., Li, B., and Li, L.
\newblock Follow your path: a progressive method for knowledge distillation.
\newblock In \emph{Machine Learning and Knowledge Discovery in Databases.
  Research Track: European Conference, ECML PKDD 2021, Bilbao, Spain, September
  13--17, 2021, Proceedings, Part III 21}, pp.\  596--611. Springer, 2021.

\bibitem[Socher et~al.(2013)Socher, Perelygin, Wu, Chuang, Manning, Ng, and
  Potts]{sst2013}
Socher, R., Perelygin, A., Wu, J., Chuang, J., Manning, C.~D., Ng, A., and
  Potts, C.
\newblock Recursive deep models for semantic compositionality over a sentiment
  treebank.
\newblock In \emph{Proceedings of the 2013 Conference on Empirical Methods in
  Natural Language Processing}, pp.\  1631--1642, Seattle, Washington, USA,
  2013. Association for Computational Linguistics.

\bibitem[Sun et~al.(2019)Sun, Cheng, Gan, and Liu]{sun2019patient}
Sun, S., Cheng, Y., Gan, Z., and Liu, J.
\newblock Patient knowledge distillation for {BERT} model compression.
\newblock In \emph{Proceedings of the 2019 Conference on Empirical Methods in
  Natural Language Processing and the 9th International Joint Conference on
  Natural Language Processing (EMNLP-IJCNLP)}, pp.\  4323--4332, Hong Kong,
  China, 2019. Association for Computational Linguistics.
\newblock \doi{10.18653/v1/D19-1441}.

\bibitem[Sun et~al.(2020{\natexlab{a}})Sun, Gan, Fang, Cheng, Wang, and
  Liu]{sun2020contrastive}
Sun, S., Gan, Z., Fang, Y., Cheng, Y., Wang, S., and Liu, J.
\newblock Contrastive distillation on intermediate representations for language
  model compression.
\newblock In \emph{Proceedings of the 2020 Conference on Empirical Methods in
  Natural Language Processing (EMNLP)}, pp.\  498--508, Online,
  2020{\natexlab{a}}. Association for Computational Linguistics.
\newblock \doi{10.18653/v1/2020.emnlp-main.36}.

\bibitem[Sun et~al.(2020{\natexlab{b}})Sun, Yu, Song, Liu, Yang, and
  Zhou]{sun2020mobilebert}
Sun, Z., Yu, H., Song, X., Liu, R., Yang, Y., and Zhou, D.
\newblock {M}obile{BERT}: a compact task-agnostic {BERT} for resource-limited
  devices.
\newblock In \emph{Proceedings of the 58th Annual Meeting of the Association
  for Computational Linguistics}, pp.\  2158--2170, Online, 2020{\natexlab{b}}.
  Association for Computational Linguistics.
\newblock \doi{10.18653/v1/2020.acl-main.195}.

\bibitem[Taori et~al.(2023)Taori, Gulrajani, Zhang, Dubois, Li, Guestrin,
  Liang, and Hashimoto]{alpaca}
Taori, R., Gulrajani, I., Zhang, T., Dubois, Y., Li, X., Guestrin, C., Liang,
  P., and Hashimoto, T.~B.
\newblock Stanford alpaca: An instruction-following llama model.
\newblock \url{https://github.com/tatsu-lab/stanford_alpaca}, 2023.

\bibitem[Trinh \& Le(2018)Trinh and Le]{trinh2018simple}
Trinh, T.~H. and Le, Q.~V.
\newblock A simple method for commonsense reasoning.
\newblock \emph{arXiv preprint arXiv:1806.02847}, 2018.

\bibitem[Vaswani et~al.(2017)Vaswani, Shazeer, Parmar, Uszkoreit, Jones, Gomez,
  Kaiser, and Polosukhin]{vaswani2017attention}
Vaswani, A., Shazeer, N., Parmar, N., Uszkoreit, J., Jones, L., Gomez, A.~N.,
  Kaiser, L., and Polosukhin, I.
\newblock Attention is all you need.
\newblock In Guyon, I., von Luxburg, U., Bengio, S., Wallach, H.~M., Fergus,
  R., Vishwanathan, S. V.~N., and Garnett, R. (eds.), \emph{Advances in Neural
  Information Processing Systems 30: Annual Conference on Neural Information
  Processing Systems 2017, December 4-9, 2017, Long Beach, CA, {USA}}, pp.\
  5998--6008, 2017.

\bibitem[Wang et~al.(2019)Wang, Singh, Michael, Hill, Levy, and
  Bowman]{wang2018glue}
Wang, A., Singh, A., Michael, J., Hill, F., Levy, O., and Bowman, S.~R.
\newblock {GLUE:} {A} multi-task benchmark and analysis platform for natural
  language understanding.
\newblock In \emph{7th International Conference on Learning Representations,
  {ICLR} 2019, New Orleans, LA, USA, May 6-9, 2019}. OpenReview.net, 2019.

\bibitem[Wang et~al.(2020)Wang, Wei, Dong, Bao, Yang, and Zhou]{wang2020minilm}
Wang, W., Wei, F., Dong, L., Bao, H., Yang, N., and Zhou, M.
\newblock Minilm: Deep self-attention distillation for task-agnostic
  compression of pre-trained transformers.
\newblock \emph{arXiv preprint arXiv:2002.10957}, 2020.

\bibitem[Wang et~al.(2021)Wang, Bao, Huang, Dong, and Wei]{wang2020minilmv2}
Wang, W., Bao, H., Huang, S., Dong, L., and Wei, F.
\newblock {M}ini{LM}v2: Multi-head self-attention relation distillation for
  compressing pretrained transformers.
\newblock In \emph{Findings of the Association for Computational Linguistics:
  ACL-IJCNLP 2021}, pp.\  2140--2151, Online, 2021. Association for
  Computational Linguistics.
\newblock \doi{10.18653/v1/2021.findings-acl.188}.

\bibitem[Warstadt et~al.(2019)Warstadt, Singh, and Bowman]{cola2018}
Warstadt, A., Singh, A., and Bowman, S.~R.
\newblock Neural network acceptability judgments.
\newblock \emph{Transactions of the Association for Computational Linguistics},
  7:\penalty0 625--641, 2019.
\newblock \doi{10.1162/tacl_a_00290}.

\bibitem[Williams et~al.(2018)Williams, Nangia, and Bowman]{mnli2018}
Williams, A., Nangia, N., and Bowman, S.
\newblock A broad-coverage challenge corpus for sentence understanding through
  inference.
\newblock In \emph{Proceedings of the 2018 Conference of the North {A}merican
  Chapter of the Association for Computational Linguistics: Human Language
  Technologies, Volume 1 (Long Papers)}, pp.\  1112--1122, New Orleans,
  Louisiana, 2018. Association for Computational Linguistics.
\newblock \doi{10.18653/v1/N18-1101}.

\bibitem[Xu et~al.(2020)Xu, Zhou, Ge, Wei, and Zhou]{xu2020bert}
Xu, C., Zhou, W., Ge, T., Wei, F., and Zhou, M.
\newblock {BERT}-of-theseus: Compressing {BERT} by progressive module
  replacing.
\newblock In \emph{Proceedings of the 2020 Conference on Empirical Methods in
  Natural Language Processing (EMNLP)}, pp.\  7859--7869, Online, 2020.
  Association for Computational Linguistics.
\newblock \doi{10.18653/v1/2020.emnlp-main.633}.

\bibitem[Zhu et~al.(2015)Zhu, Kiros, Zemel, Salakhutdinov, Urtasun, Torralba,
  and Fidler]{zhu2015aligning}
Zhu, Y., Kiros, R., Zemel, R.~S., Salakhutdinov, R., Urtasun, R., Torralba, A.,
  and Fidler, S.
\newblock Aligning books and movies: Towards story-like visual explanations by
  watching movies and reading books.
\newblock In \emph{2015 {IEEE} International Conference on Computer Vision,
  {ICCV} 2015, Santiago, Chile, December 7-13, 2015}, pp.\  19--27. {IEEE}
  Computer Society, 2015.
\newblock \doi{10.1109/ICCV.2015.11}.

\bibitem[Zuo et~al.(2022)Zuo, Zhang, Liang, He, Zhao, and Chen]{zuo2022moebert}
Zuo, S., Zhang, Q., Liang, C., He, P., Zhao, T., and Chen, W.
\newblock Moebert: from bert to mixture-of-experts via importance-guided
  adaptation.
\newblock \emph{arXiv preprint arXiv:2204.07675}, 2022.

\end{thebibliography}
